\documentclass[journal]{elsarticle}

\usepackage{lineno,hyperref}
\modulolinenumbers[5]

\journal{Journal of Computer Speech and Language}

\usepackage[T1]{fontenc}
\usepackage[utf8]{inputenc}
\usepackage{gensymb}
\usepackage{multirow}
\usepackage{amsmath}
\usepackage{array}
\usepackage[normalem]{ulem}
\newcolumntype{$}{>{\global\let\currentrowstyle\relax}}
\newcolumntype{^}{>{\currentrowstyle}}

\usepackage{multirow}
\usepackage{pbox}
\usepackage[table]{xcolor}
\usepackage{xcolor,colortbl}

\definecolor{Gray}{gray}{0.55}
\definecolor{Gray1}{gray}{0.75}
\definecolor{SkyBlue}{rgb}{0.5,0.8,0.9}
\definecolor{DarkGray}{rgb}{0.76,0.76,0.76}
\definecolor{LightGray}{rgb}{0.91,0.91,0.91}
\definecolor{LightCyan}{rgb}{0.88,1,1}

\newcolumntype{a}{>{\columncolor{Gray}}c}
\newcolumntype{b}{>{\columncolor{white}}c}

\makeatletter
\newcommand{\thickhline}{%
	\noalign {\ifnum 0=`}\fi \hrule height 2pt
	\futurelet \reserved@a \@xhline
}
\newcolumntype{"}{@{\hskip\tabcolsep\vrule width 1pt\hskip\tabcolsep}}
\makeatother

\newcommand*\justifyme{%
	\fontdimen2\font=0.4em% interword space
	\fontdimen3\font=0.2em% interword stretch
	\fontdimen4\font=0.1em% interword shrink
	\fontdimen7\font=0.1em% extra space
	\hyphenchar\font=`\-% allowing hyphenation
}
\begin{document}

\begin{frontmatter}

%\title{Elsevier \LaTeX\ template\tnoteref{mytitlenote}}
\title{Annotating and Modeling Empathy \linebreak in Spoken Conversations}


\author{Firoj~Alam, Morena~Danieli and Giuseppe~Riccardi\corref{mycorrespondingauthor}}
\cortext[mycorrespondingauthor]{Corresponding author}
\address{Department of Information
	Engineering and Computer Science, \\University of Trento, Italy, 38123.}
\ead{\{firoj.alam, morena.danieli, giuseppe.riccardi\}@unitn.it}
\fntext[myfootnote]{\copyright~2017. This manuscript version is made available under the CC-BY-NC-ND 4.0 license http://creativecommons.org/licenses/by-nc-nd/4.0/
}

\begin{abstract}
Empathy, as defined in behavioral sciences, expresses the ability of human beings to recognize, understand and react to emotions, attitudes and beliefs of others. 
%The current state-of-art definition of empathy makes it very difficult to operationalize it in interactive scenarios, for example, spoken interaction.  
%The lack of an operational definition of empathy makes it difficult to measure it. 
In this paper, we address two related problems in automatic affective behavior analysis: the design of the annotation protocol and the automatic recognition of empathy from human-human dyadic spoken conversations. We propose and evaluate an annotation scheme for empathy inspired by the {\em modal} model of emotions. The annotation scheme was evaluated on a corpus of real-life, dyadic spoken conversations. In the context of behavioral analysis, we designed an automatic segmentation and classification system for empathy. Given the different speech and language levels of representation where empathy may be communicated, we investigated features derived from the lexical and acoustic spaces.
The feature development process was designed to support both the fusion and automatic selection of relevant features from a high dimensional space. The automatic classification system was evaluated on call center conversations where it showed significantly better performance than the baseline.
\end{abstract}


\end{frontmatter}

%\linenumbers

\section{Introduction} %%%%%%%% Firoj and Morena %%%%%%%%
\label{sec:intro}
%%%%%%%%%%%%%%%% \textbf{ What is empathy } %%%%%%%%%%%%%%%

Research in human emotions is an interdisciplinary field of study with relevant contributions from different scientific fields including neuroscience, psychology, and more recently computer science. In the latter discipline, the motivations for studying human emotions are driven by the goal of training virtual agents and improving human-computer interaction. Training intelligent agents to understand human emotions poses fundamental research challenges in \textbf{a)} the annotation of emotion examples for computers to learn from, \textbf{b)} the design of algorithms to automatically recognize and \textbf{c)} synthesize emotions. For example, in human-machine dialog research there has been growing interest in how to coordinate linguistic and paralinguistic signals and their impact on the performance of dialogs \cite{riccardi2005grounding}. Progress has been reported in modeling full spectrum of human short and long-term states such as mood and emotion, and in modeling personality traits \cite{alam2013comparative} by analyzing paralinguistic phenomena \cite{schuller2009interspeech}, facial expressions, gestures \cite{zeng2009survey} and bio-signals \cite{wagner2005physiological}. Despite this growing interest there are  very few studies that explicitly investigate complex emotions such as empathy. 

The case for empathy is compelling because the ability to recognize and reproduce empathic behaviour of conversants would greatly improve the applicability of virtual agents. 
The concept of  \textit{empathy} was investigated in experimental psychology starting at the beginning of last century, when Titchener in $1909$ \cite{titchener1909lectures} coined the very word {\em empathy} as a translation of the German term {\em Einf{\"u}hlung}. Since then this concept has been widely used to refer to a range of pro-social emotional behaviors, from sympathy to compassion, including accurate understanding of the other person's feelings. Recently, the hypothesis that one's empathy is triggered by understanding and sharing  others' emotional states has found neuroscientific underpinnings in the discovery of the mirror-neurons system. This system is hypothesized to embody unconscious routines of emotional and empathic behaviors in interpersonal relations. These include action understanding, attribution of intentions (\textit{mind-reading}), and recognition of emotions and sensations \cite{gallese1998mirror}. In everyday life, empathy supports important aspects of inter-personal communication to an extent where some psychic diseases that affect the relationship with other persons, such as autism and Asperger syndrome, are explained in terms of  impairment of empathic ability \cite{baron2013understanding}.
In computer science literature, the authors in~\cite{boukricha2013computational} evaluate the synthesis of empathic virtual agents, while ~\cite{xiao2012analyzing}  recognize the empathy of a therapist during interactions with his patients, and ~\cite{kumano2011analyzing} estimate the empathic behavior in meeting conversations. 

Our interest in the problem of empathy is driven by the goal of learning computational models of complex emotional states on a psychologically motivated model of human emotions. In this paper, we refer to the psychological definition of empathy by Hoffman, who defines it as ``an emotional state triggered by another's emotional state or situation, in which one feels what the other feels or would normally be expected to feel in his situation'' (\cite{hoffman2008empathy}). In particular, we focus on identifying the instances of empathy emerging in spoken real-life conversations. 
%In order to develop a computational model of empathy we need to design an experiment where the unfolding of emotions occurring in an empathic process can be observed. 
To develop a computational model of empathy, we designed an experimental paradigm to observe the unfolding of emotions in the context of empathic behavior.
%To this end 
For that we adopt the {\em modal} model of emotions by Gross \cite{gross1998emerging} as a promising framework for defining an \textit{operational} concept of empathy. We use this framework for developing the annotation guidelines for the annotation of empathy in a real-life spoken conversation corpus. Our research is organized into three phases: data observation and analysis, corpus annotation, and automatic classification. The guidelines for the annotation of the spoken corpus recommend a continuous selection of the emotion perception on the speech channel, and the assignment of discrete labels.

The main contributions of this paper are 1) the design and evaluation of an annotation model of empathy in spoken dialogues,  2) the training of an automatic classification system for empathy based on acoustic, lexical and psycholinguistic features\footnote{Psycholinguistic features are extracted from transcriptions. It is a knowledge-based approach, in which a lexicon is used to compute the correlation between words and psycholinguistic categories. More details can be found in Section \ref{sssec:psycholinguistic_features}.}, and 3) the evaluation of the classification system on a corpus of human-human spoken conversations. To the best of our knowledge, this is the first published research on perceiving, annotating and automatically recognizing empathy from human-human spoken conversations in real-life interactions. 

The paper is organized as follows. In Section \ref{sec:background-research}, we provide a review of relevant research on empathy. We then briefly discuss the modal model of emotions and its relevance to our work in Section \ref{sec:modal-model}. In Section \ref{sec:operational-model}, we describe the proposed annotation scheme for the annotators. In Section \ref{sec:corpus-analysis}, we provide the corpus analysis and investigate the speech correlates of empathy. In Section \ref{sec:classification}, we describe the computational architecture of the automatic segmentation and classification system. We then provide the results and discussion of our experiments in Section \ref{ssec:results_discussion}. We discuss the research challenges and future work in Section \ref{ssec:limitations_future_work}.

\section{Background Research}
\label{sec:background-research}

\subsection{Psychology and Neuroscience Research} 
\label{ssec:psychology-research}
Over the past decades there have been significant efforts in investigating empathy in the fields of psychology and neuroscience \cite{zakireview2012} \cite{calvo2010affect}. The complexity of the neural and psychological phenomena to be accounted for is huge and, in part, that complexity explains the existence of several psychological definitions of empathy. For example, the work in \cite{preston2012many} accounts for  different empathic phenomena occurring in the literature on empathy, and \cite{batson2009these} examines eight distinct phenomena commonly labeled as empathy including  emotional contagion, sympathy, projection, and affective inferential processes. Decety and Lamm \cite{decety2006human} observe that some of the different definitions of empathy may share the underlying concept of  ``[...] an emotional experience that is more congruent with another’s situation than with one’s own''. The authors also state that empathic emotional experiences imply self-other differentiation, as opposed to emotional contagion. Among the different definitions of psychological theory there are some common features.
%Actually if we abstract from the differences in the theoretical perspectives, we  may find some common features.
Most of the definitions describe empathy as a type of emotional experience and/or emotional state. Moreover, the different definitions can be divided into two main classes. One encompasses the cognitive aspects of empathic behavior, such as one's ability to accurately understand the feelings of another person. The other class entails  sharing or the subject's internal mimic of such feelings such as sympathy, emotional contagion, and compassion. The work in \cite{zaki2014empathy} takes a different perspective on empathy by focusing on the role of motivations for explaining a particular feature of the empathic experience, i.e., its oscillation between automaticity and dependency from the situational context.

\subsection{Computational Models}  
\label{ssec:computational-research}

\subsubsection{Research in Speech Communication}  
\label{sssec:computational-research}

Computational models of emotional states are needed to design machines that can understand and interact {\em affectively} with humans. Different signal components have been considered for analyzing the emotional instances in speech. Both verbal and non-verbal levels of spoken communication \cite{gesn1999development} have been considered since both are suggested to embody the expressive potential of language. Major focus has been devoted to the paralinguistic features of  emotional speech, on the basis of the experimental evidence that emotional information is mostly conveyed by those levels (see \cite{schuller2011recognising} for a state-of-the-art review).

In the field of spoken language processing there are several collections of emotional annotated corpora. The authors in~\cite{el2011survey} report a significant disparity among those corpora in terms of complexity of the annotated emotions, explicitness of the emotion definitions, and identification of the annotation units. Most emotional corpora have been designed to perform specific tasks such as emotion recognition or emotional speech synthesis \cite{tesser2004modelli, zovato2004prosodic}. The associated annotation schemes often depend on the specific tasks as well. The HUMAINE project \cite{douglas2003description} and the emotion multilingual collection in \cite{origlia2010automatic} base their annotation schemes on sets of discrete emotional lexical items or basic emotions.

Providing explicative models for annotating the emotional process itself in naturally occurring conversations is still an open challenge. Efforts in this direction are currently being made in the affective computing research, where awareness about the need for continuous annotation is  increasing. The models that foster this approach, such as those discussed in \cite{mckeown2012semaine} and \cite{ringeval2013introducing}, require  annotators to continuously assign values to emotional dimensions and sets of emotional descriptors. Metallinou and Narayanan \cite{metallinou2013annotation} emphasize that continuous annotation has several benefits, as well as some open challenges. One interesting finding is that continuous annotation may show regions which are characterized by perceived transitions of the emotional state. In~\cite{metallinou2013annotation}, authors report a high number of factors that may affect inter-annotator agreement, such as person-specific annotation delays and confidence in understanding emotional attributes. 
There are very few studies in terms of empathy classification and most of them are carried out within controlled scenarios. Kumano et al. \cite{kumano2011analyzing} studied four-party meeting conversations to estimate and classify empathy, antipathy and unconcerned emotional interactions utilizing facial expression, gaze and speech-silence features. In \cite{xiao2012analyzing} and \cite{xiao2014modeling}, Xiao et al. analyzed therapists' conversation to automatically classify empathic and non-empathic utterances using linguistic and acoustic information. 

%%%%%
% Research on Sentiment Analysis
%
%%%%%%
\subsubsection{Research on Sentiment Analysis from Text}  
\label{sssec:computational-research-nlp}
%\cite{pak2010twitter}
When it comes to the term ``Sentiment Analysis'' current state-of-art research are mostly focused on extracting it from textual information \cite{pang2008opinion}. It could be from movie-reviews \cite{pang2002thumbs}, tweets \cite{bollen.al11b,paltoglou2010study}, and newspaper articles and comments \cite{celli2016mood}.   
Sentiment analysis from text has been mostly used in terms of positive and negative polarity \cite{pak.paroubek10,kouloumpis2011twitter,cambria2016senticnet,akkaya2009subjectivity,paltoglou2010study}. More detailed dimensions are proven to be very useful. For example, moods such as tension, depression, anger, vigor, fatigue, and confusion in tweets have been found to be good predictors of stock market exchanges \cite{bollen.al11b}. It has also been found that it is possible to predict anger, sadness, and joy from LiveJournal blogs \cite{nguyen2010classification}. A study also suggests that it is also possible to distinguish Twitter users who are likely to share content generating joy or amusement from the ones who are likely to share content generating sadness, anger or disappointment with an accuracy of around 61\% \cite{celli2016mood}. 

A document contains different subjects/topics where each topic is associated with a polarity. In \cite{nasukawa2003sentiment}, Nasukawa and Yi present that semantic information and sentiment lexicon can capture such polarity associated with the topic. Pand and Lee \cite{pang2004sentimental} present an approach, which first detects the subjectivity portion of text using a graph-based method then use a supervised classifier to detect the polarity. 
Entity based polarity detection first detect (e.g., people, places, things) from text then assigns scores indicating positive or negative opinion toward that. The study of Godbole in \cite{godbole2007large}, present a system that captures sentiment of news and blog entities. 
%\cite{pang2008opinion}
Lin and He \cite{lin2009joint} proposed a joint sentiment/topic model by using Latent Dirichlet Allocation (LDA) with a sentiment layer, which detects topics of a document and assigns sentiment polarity on the topics. 
In \cite{taboada2011lexicon}, Taboada et al. present a lexicon-based approach to assign a positive or negative label to a text that captures the text's opinion towards its main subject matter. 
Maas et al. \cite{maas2011learning} proposed a probabilistic model of documents, which learns word representations and used LDA to learns topics and used logistic regression for the classification. 
The study of Kouloumpis et al. in \cite{kouloumpis2011twitter} present the use of n-gram, parts-of-speech tags and some lexical features for classifying tweets with sentiment polarity. 
In \cite{cambria2016affective}, the authors present a hybrid framework for sentiment analysis that includes a knowledge-based system and a machine learning module. A research application SentiStrength \cite{thelwall2010sentiment} utilizes a different source of information to assign a sentiment score to a short text \cite{thelwall2011sentiment,stieglitz2013emotions}. Such information includes word-list of sentiment, idioms, emoticons, negating words, linguistic rules and sentiment polarity classification algorithms.
Last, but not least, a survey of sentiment analysis can be found in \cite{liu2012survey}.

\section{The Modal Model of Emotions}
\label{sec:modal-model}
\begin{figure}[h]
\centering
\includegraphics[width=3.5in]{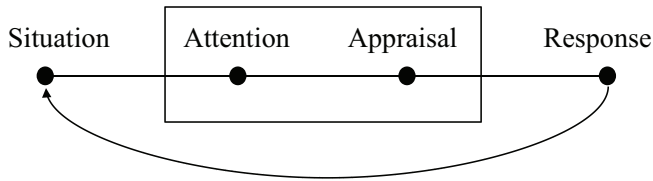}
\caption{The modal model of emotion \cite{gross1998emerging}.}
\label{fig:modal_model}
%\vspace*{-4mm}
\end{figure}

Many psychologists have studied emotional episodes from the point of view of appraisal dimensions. Gross ~\cite{gross1998emerging} has provided evidence that concepts such as \textit{emergence} --- derivation from the expectations of relationships --- and \textit{unfolding} --- sequences that persist over time --- may help in explaining emotional events. It has been shown that temporal unfolding of emotions can be conceptualized and experimentally tested \cite{sander2005systems}. The modal model of emotions developed by Gross \cite{gross1998emerging, gross2007emotion} emphasizes the attentional and appraisal acts underlying the emotion-arousing process. In Figure \ref{fig:modal_model}, we provide the original schema of Gross model. The individuals' core {\em Attention-Appraisal} processes (included in the box) are affected by the {\em Situation}  that is defined objectively in terms of physical or virtual spaces and objects. The \textbf{Situation} compels the \textbf{Attention} of the individual; it triggers an \textbf{Appraisal} process and gives rise to coordinated and malleable \textbf{Responses}. It is important to note that this model is dynamic and the situation may be modified (directed arc from the \textbf{Response} to the \textbf{Situation}) by the actual value of the \textbf{Response} generated by the {\em Attention-Appraisal} process. 
%In this model, emotions are seen as a way of experiencing the world: they are distinct functional states \cite{gross2011emotion}, and the appraisal acts describe the content of those functional states within a context. 
The modal model of emotions provides a useful framework for describing the contextual dynamics of emotions within an {\em affective scene}(\cite{danieli2015emotion}), since it decomposes the emotional process in terms of situation selection, attentional deployment, and situational modification. Gross' informal model can provide insight in the process where the empathic responses may modify the initial emotional context.  
% MORENA : changes have been made in this paragraph
For guiding the observers through their perception process we rely on the modal model of emotions. That framework may provide the observers with grounded descriptions of the unfolding of the emotions in the affective scene. The annotator will need to identify the emergence, appraisal, and effects of emotions felt by the call center agent in the ongoing (sub-)dialogs. In doing so, the annotator should also be able to perceive if, and to what extent, an empathic response may modify a situation where other emotions such as frustration or anger are expressed by the customer. In section~\ref{ssec:operational_definition} we describe how the modal model of emotions applies to our annotation model of empathy. 

%MORENA: I have done changes in this section
\section{The Empathy Annotation Model}
\label{sec:operational-model}
The goals of the annotation model for empathy is to provide empirical evidence of the reference definition and annotated signal samples to train a computational model. In the following sections we describe some challenges in annotating real-life stimuli (section \ref{ssec:annotation_unit}), we report on the qualitative data analysis of the dialogue corpus (section \ref{ssec:qualitative_data_analysis}), we introduce the operational definition of empathy (section~\ref{ssec:operational_definition}) in the context of call center conversations, and we describe the annotation process (section \ref{ssec:annotation_procedure}) and the evaluation of the annotators' decisions (section \ref{ssec:evaluation_annotation_model}).  

\subsection{Annotation of Real-Life Spoken Conversations} %%%%%%% Morena %%%%%%%%%%
\label{ssec:annotation_unit}

The annotation unit is the stimulus presented to the observer (annotator) to perform a selection task over a decision space such as the set of emotional state tags. In general the annotator may be presented with images or speech segments (stimuli), and a set of emotional labels. The stimulus is defined in terms of the medium the emotion is being {\em transmitted} through and its content and context. The medium may be speech \cite{liscombe2005using}, image \cite{pantic2000automatic} or multimodal \cite{metallinou2012context}. The content refers to the information encoded in the stimulus signal such as facial expression of anger or a speech utterance. The context of the stimulus is represented by the spatial or temporal signals neighboring the stimulus. Knowledge of the context may be crucial in interpreting the cause of emotion manifestations. For a speech stimulus in a conversation the context is represented by the preceding dialog turns \cite{liscombe2005using}. The above description of the annotation unit does not assume a univocal relationship between the occurrence of an emotion and its corresponding expression.

Most research in affective computing has been focused on stimuli that are designed in advance and are artificially generated. Respective examples of such stimuli are sentences to be read and actors enacting affective scenes \cite{cowie2001emotion}. Another limitation of acted annotation tasks is that stimuli are {\em context-free} and speech utterances or images are annotated in isolation. 

The limitations of determining the temporal boundaries of the annotation units have motivated researchers to investigate the process of continuous annotation \cite{metallinou2013annotation, mckeown2012semaine}. Yet, state-of-the-art complete continuous affective scene annotation techniques are highly demanding for observers and are not very effective in terms of inter-annotator agreement. 
%However, complete continuous annotation of the affective scene is very demanding in terms of observer's selection task and at the same time not very effective (in terms of inter-annotators' agreement).
In our work we address the task of defining and searching the annotation unit in non-acted, real-life spoken conversations. We describe below how the annotation model we propose exploits the conversation context for the perception and annotation of the empathic events.

\subsection{Qualitative Data Analysis}
\label{ssec:qualitative_data_analysis}

We have performed qualitative analysis of the spoken conversation corpus to gain insights into the speech signal instantiations of emotions and empathy. The analysis was carried out over a corpus of human-human dyadic Italian call center conversations that will be described in Section \ref{sec:corpus-analysis}. We analyzed one hundred conversations (more than $11$ hours), randomly selected from the corpus, in particular the dialog turns where the speech signal showed the emergence of emotions (e.g. frustration, anger) on the customer channel and empathy on the agent channel. 

The outcome of the qualitative analysis has supported the view that emotionally relevant conversational segments are often characterized by significant transitions in the paralinguistic patterns or the occurrence of lexical cues. As expected, such variations may co-occur not only with emotionally-connoted words but also with functional parts of speech (POS) such as Adverbs and Interjections. Phrases and Verbs, as shown in Table \ref{table:annotation_example}, could also lexically support the expression of emotional states. 
%We evaluated the communicative situation in terms of appraisal of the transition from a neutral to an emotionally-connoted state. 

In Table \ref{table:annotation_example}, we present a dialog excerpt with annotations. The dialog excerpt is reported in the first column of the table, where {\bf C} is the customer, and {\bf A} is the agent. The situation is the following: {\bf C} is calling because a payment is actually overdue: he is ashamed for not being able to pay immediately and his speech has plenty of hesitations. This causes an empathic response by {\bf A}: that emerges from the intonation profile of {\bf A}'s reply and from her lexical choices. In the second question of {\bf A}'s turn, she uses the hortatory first person plural instead of the singular one. Also, the rhetorical structure of {\bf A}'s turn, i.e., the use of questions instead of assertions, conveys her empathic attitude. The annotator identified the context, i.e., the preceding dialog turn, and perceived the intonation variation. The annotator could then mark the speech segment corresponding to the intonation unit starting with the word {\em proviamo} ({\em let us try}) as onset of the emotional process.

\begin{table}[!t]
\caption{An excerpt from a telephone conversation where  the agent ({\bf A}) is empathic towards a customer ({\bf C}). The agent perceives the customer's feeling and proactively takes actions to cope with customer's emotional discomfort. English translations are in italics.}
\label{table:annotation_example}
\centering
\begin{tabular}{|c"c|}

\rowcolor{black!30} \bfseries Dialog excerpt & \bfseries Notes \\ \hline
\pbox{7cm}{
	{\bf C:} Ascolti ... io ho una fattura scaduta di 833 euro vorrei sapere ... tempo in cui posso pagarla.\\
	\textit{(Listen... I have an 833 euros overdue bill... I would like to know... the time left to pay it.)}\\
	{\bf A:} Ma perché non ha chiesto il rateizzo di questa fattura? Proviamo a far il rateizzo, ok? Così gliela blocco e lei ha più tempo per effettuare il pagamento.\\
	\textit{(Why did not you ask to pay it in installments? We try to divide it into installments, is it ok for you? So I stop the overdue notices and you will have more time to pay)}
}&
\pbox{4cm}{\fontfamily{pcr}\ttfamily\scriptsize\slshape{\justifyme The tone of voice and the hesitations of the customer show that she is not angry, she is ashamed for not being able to pay immediately the bill. This causes an empathic reply in the Operator’s attitude.} \\
	\fontfamily{pcr}\ttfamily\scriptsize\slshape{\justifyme The selection of the speech act (question instead of authoritative declarative), the rhetorical structure of the second question, the lexical choice of ``proviamo'', instead of - for instance, ``adesso provo a vedere...'', all these contribute to prevent the customer's feeling of being inadequate or ashamed. }} \\ \hline
\end{tabular}
%\vspace*{-6mm}
\end{table}

\subsection{The Operational Definition of Empathy}
\label{ssec:operational_definition}

Following the modal model of emotions and the annotation model discussed above, we first describe the context of the situation, the attention, the appraisal and the response components of the empathic process. 

\textbf{The context of the situation:} In call center conversations, customers may call to ask for information or for help to resolve technical or accounting issues. Agents are supposed to be cooperative and empathic and they are trained for the task. However, variabilities in the agents' or customers' personalities, behaviors and random events lead to statistical variations in the emotional unfolding. 
% of affective scenes.
The operational definition of empathy  that may be applied to this context requires the annotators to be informed of the social context and task. They are trained to focus over sub-dialogues where the agent anticipates or proposes solutions and clarifications (\textbf{attention}), based on the understanding of the customer's problem (\textbf{appraisal}). As a consequence, the acts of the agent may prevent or releave customer's unpleasant feelings (\textbf{response}). Therefore, we operationally define empathy as \textit{a situation where an agent anticipates or views solutions and clarifications, based on the understanding of a customer's problem or issue, that can relieve or prevent the customer's unpleasant feelings}.  

The selection of the stimuli includes a continuous search of the speech segments preceding and following the perception of the 
% {\em onset} 
outset of the 
% empathic event.
empathic segment. 
The task of the annotator is to identify the context (left of the 
%onset) 
outset) and the target (right of the 
%onset) 
outset) empathy segment. 
The context is defined to be \textit{neutral} with respect to the target empathic segment. This does not mean that emotions rather than empathy cannot be manifested by the speaker in the speech segment that the annotator identifies as context, but that those possibly occurring emotions are not recognized as empathy by the annotator. The reference to a neutral, i.e., \textit{non-empathic}, segment supports the annotators in their perception process  while identifying the outset of the empathic segment. 
This annotation protocol was applied to annotate sequences of basic emotions occurring in affective scenes (\cite{morena2015}).

Instructions were given to the annotators to achieve the most confident decision in identifying and tagging the empathy segments, based on the perceived paralinguistic and/or linguistic cues. Based on prior research \cite{thompson2003perceiving}, we advised them to focus their attention on the variations in the paralinguistic cues (e.g., pitch rise or fall). 
The annotation guidelines provided examples like the ones described in section~\ref{ssec:qualitative_data_analysis}.

\subsection{The Annotation Procedure}
\label{ssec:annotation_procedure}

The typical corpus annotation process starts with the selection of a set of conversations and then the annotation of the speech segments in each conversation. In this work, we focused on first instances of empathy segments within each conversation. The rationale behind this choice was to maximize the number of annotated conversations with many speakers given limited resources.  
%In the annotation process, we set the goal of maximizing the number of annotated conversations. For this reason, we annotated the first instance of neutral-empathy segment pairs within each conversation. 
Annotators manually refined the boundaries of the segments generated by an off-the-shelf Speech/Non-Speech segmenter.
The annotators tagged the {\tt Empathy} segments on the agent channel and the basic emotions ({\tt Frustation} and {\tt Anger}) on the customer channel. The reason of choosing these basic emotions was due to the limited resources for annotation. Another reason was that we observed these emotions most frequently occurred in our corpus. 
%these are the most frequently occuring cases in our corpus.
Annotators were instructed to select the candidate segment pairs with a decision refinement process. Once the relevant speech region was identified, the annotators could listen to the speech as many times as they needed to judge if the selected segment(s) could be tagged with any of the target labels. The average per-conversation annotation time was 18 minutes for an average duration of a conversation of 6 minutes.
Two annotators completed the annotation task. They used the EXMARaLDA Partitur Editor \cite{schmidt2004transcribing}. The corpus has been annotated with four labels: {\tt Empathy}, {\tt Anger}, {\tt Frustration}, {\tt Neutral} to describe the complete affective scene of the context (more in Section \ref{sec:corpus-analysis}).

\subsection{Evaluation of the Annotation Model}  %%%%%%% Morena %%%%%%%%%%
\label{ssec:evaluation_annotation_model}

To assess the reliability of the annotation model, we designed the following evaluation task. Two annotators, with psychology background, worked independently over a set of 64 spoken conversations (approximately $7$ hours of speech) randomly selected from the call center corpus. The annotators were of similar age, same ethnicity and opposite gender. We intended to assess if the annotators could perceive a change in the emotional state at the same signal position ({\tt Neutral} leading into {\tt Empathy}), as well as their agreement with the assignment of the empathy label. In case of disagreement between the annotators a consensus meeting tried to reach a consensus. 
%In 53.1\% of the annotated segments the two annotators perceived the empathic attitude of the agent in  adjacent segments, while  31.2\% of the speech segments were tagged with empathy labels on the same onset position.
The inter-annotator agreement for the annotated segments between the two annotators is $53.1\%$, where the agreement on the same onset position is $31.2\%$. Annotating in continuous time space is typically a difficult task. Moreover, when it comes to annotate emotional aspects it became much more difficult due to the perception of the annotators in defining the emotional segment boundary in time space. This difficulty results in the low agreement of emotional onset boundary detection, which is reasonable.  

To measure the reliability of the annotations  we calculated inter-annotator agreement by using the kappa statistics
\cite{carletta1996assessing}. It is frequently used to assess the degree of agreement among any number of annotators by excluding the probability that they agree by chance. The kappa coefficient ranges between 0 (agreement is due to chance) and 1 (perfect agreement). Values above $0.6$ suggest acceptable agreement (\cite{krippendorf1980} \cite{landiskoch1977agreement},\cite{calleias2009agreement}). Our annotation task was challenging because it combined categorical annotation with continuous perception of the slowly varying emotion expression from speech-only stimuli. Thus we evaluated the agreement between annotators based on a partial match: two  annotators agreed on the selection of the onset time stamps within a tolerance window of 5 sec. We found reliable results with kappa value for empathy $0.74$. 

Most categorical emotion annotation research in speech deals with lower human agreement (greater than $0.50$) maybe due to a variety of factors, including short audio clips or utterance (\cite{abrilian2005emotv1, liscombe2005using}), multi-label annotation tasks, and annotator agreement when the annotation task is based on continuous and discrete label annotations \cite{metallinou2013annotation}. In our case the positive evaluation results may be motivated by the operationalized definition of empathy, by the observability of the complete paralinguistic and linguistic contexts, and finally by the binary annotation task.

\section{Corpus Analysis} 
\label{sec:corpus-analysis}
In this Section, we describe the spoken conversation corpus and its annotation statistics. We then report a correlation study and discover speech correlates in the speech signal (Section \ref{ssec:acoustic_feature_analysis}) and lexical features ( \ref{ssec:lexical_feature_analysis} ). These correlates are then used to guide the training of the automatic empathy classification model. 

\subsection{Corpus Description} 
\label{ssec:corpus-desc}

The corpus includes $1894$ randomly selected customer-agent conversations, which were collected over the course of six-months, amounting to $210$ hours of speech data. These conversations were recorded on two separate channels of $16$ bits per sample and  $8kHz$ sampling rate. The average length of the conversations was $406$ seconds.

\begin{figure}[h]
	\centering
	\includegraphics[width=4.0in]{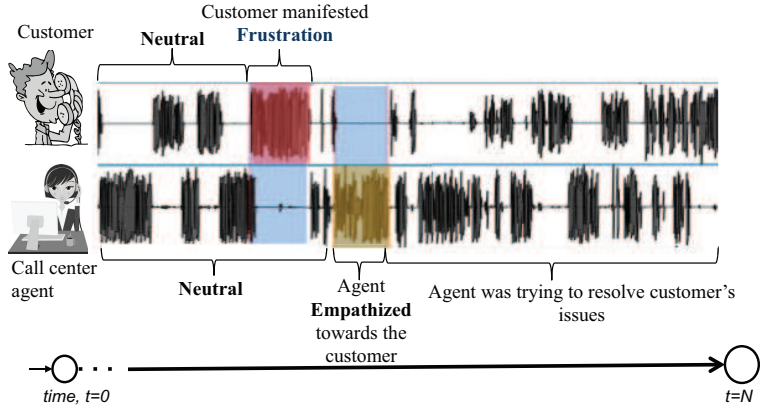}
	\caption{A part of the annotated example of an interaction between call centers agent and customer, which depicts the emotional sequence of the conversation. %From time $238.8$ to $262.4$ seconds 
    In this conversation, customer first manifested frustration, 
    %$271.7$ to $292.0$ 
    then agent empathized towards the customer and was trying to resolve customer's issues.
    %$416.8$ to $423.2$ seconds 
    }
	\label{fig:example_agent_customer}
	%\vspace*{-4mm}
\end{figure}

We analyzed the distribution of emotion state label sequences describing the affective scene in the conversations. In Figure \ref{fig:example_agent_customer}, we present an example of an emotional sequence, which manifests the affective scene. We analyzed such sequential emotional patterns in the corpus to understand the insights of the agent and customer affective behavioral aspects. More details of such findings can be found in \cite{alam2017affectivescene}. We observed that empathy was perceived by annotators in $27.72\%$ of the conversations out of the whole set as can be seen in Table \ref{table:emo-label-coo-matrix}, in which column represents agent's channel and row represents customer's channel. By analyzing only empathic conversations ($27.72\%$), we observed that when agent manifested empathy no frustration or anger was manifested by the customer in $70\%$ of the cases. This affective scene may be explained, in part, by specific training of agents in anticipating customers' issues and taking appropriate actions. For the remaining $30\%$ of the conversations customers had manifested anger or frustration or both. 

Moreover, for $12.99\%$ (see in Table \ref{table:emo-label-coo-matrix}) of the conversations out of whole set, customers  had manifested anger or frustration or both while no empathy had been perceived on the agent side. From our analysis we can then characterize the stereotypical situations that occur in the affective scene with the two following scenarios:
\begin{itemize}
\item Customer shows anger and/or frustration and the agent does not react to the speech and language cues of the customer. This is the case in which agent was expected to be empathic, but failed to recognize or react to customer's signals. 
\item Agent is empathic in response to or in anticipation of the customer's anger, frustration or both. 
\end{itemize} 

\begin{table}[h]
\centering
\caption{Co-occurrence matrix between {\tt Empathy} or {\tt Neutral} on the agent channel (columns) and {\tt Anger} or {\tt Frustration}  on the customer channel (rows). }
\label{table:emo-label-coo-matrix}
\begin{tabular}{|l|r|r|}
\hline
\multicolumn{1}{|c|}{} & \multicolumn{1}{c|}{\textbf{ \tt Empathy (\%)}} & \multicolumn{1}{c|}{\textbf{ \tt Neutral(\%)}} \\ \hline
\textbf{{\tt Anger} or {\tt Frustration}} & 156 (8.24) & 246 (12.99) \\ \hline
\textbf{\tt Neutral} & 369 (19.48) & 1123 (59.29) \\ \hline
\textbf{Total} & 525 (27.72) & 1369 (72.28) \\ \hline
\end{tabular}
\end{table}

%%%%%%%% Firoj %%%%%%%%%%
\subsection{Acoustic Feature Analysis} 
\label{ssec:acoustic_feature_analysis}

\begin{figure}[!t]
\centering
\includegraphics[width=4in]{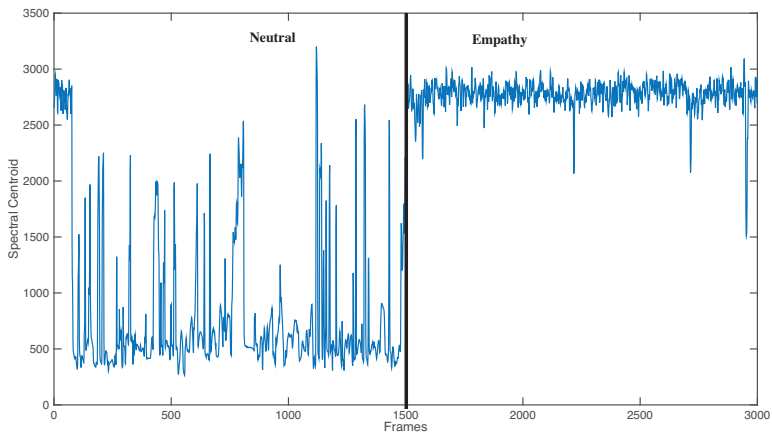}
\caption{Spectral centroid (in Hz) of an annotation unit. The onset is marked in bold. The neutral (left) segment is the context support preceding the right segment annotated with the empathy label (in this case a female speaker). Both segments are 15 seconds long.}
\label{fig:female_spectral_centroid}
\end{figure}

We investigated and compared the pattern sequences of low-level acoustic features before and after the onset point, from the neutral to the empathy segment. An example of the speech segment annotation is shown in Figure \ref{fig:female_spectral_centroid}, where we plot the spectral centroid\footnote{Spectral centroid measure the brightness (i.e., high frequency signals) of a sound. It is defined as the frequency-weighted sum of the power spectrum normalized by its unweighted sum \cite{lerch2012introduction}. Using spectral centroid we can capture how the centroid of two signals differ.} feature values across the neutral and empathy connoted segment. Each segment is 15 seconds long and the onset is marked by a vertical bold line, which separates the left (context) and right segment annotated with empathy. From the signal trend of this feature we see that there is a distinctive profile change, which is corroborated by its high statistical significance ($p-value=4.61E-51$, $d=1.4$ and $t=-16.6$ using {\em two-tailed two-sample t-test}). 

The low-level features were extracted from both left and right segment with 100 overlapping frames per second, pre-emphasis with k=0.97, and hamming-windowing, using openSMILE \cite{eyben2013recent}. For voice quality features we used gaussian windowing function. Then, we computed averages for each segment of the corresponding conversation. In order to evaluate the relevance of each feature we applied a statistical significance test, the {\em two-tailed two-sample t-test} at p-value $=0.01$.  We analyzed $45$ low-level acoustic features from five categories: pitch ($4$), loudness ($1$), zero-crossings (1), spectral ($13$), and auditory-spectrum bands (26). In Table \ref{table:feat-analysis}, we list the acoustic features that passed the significance test for male ($40$ features) and female ($34$ features) speakers. We also report effect sizes, d, which are computed using Cohen's d, as in Equation \ref{eq:cohensd}. Effect size refers to how big or small the difference in means of two groups. 
%to the how big or small between the means of two groups. 
Here two groups refers to the neutral and empathy segments. The number of samples for this analysis is $302$ conversations, $n=302$, sample size.

\begin{equation}
\label{eq:cohensd}
\begin{split}
Cohen's\;d=\frac{m1-m2}{\sigma ^{p}}
\;\;;\;\sigma ^{p}=\sqrt{\frac{\sigma{_{1}^{2}}+\sigma_{2}^{2}}{2}} 
\end{split}
\end{equation}

where $m1$ and $m2$ are the means of two samples, and $\sigma_{1}$ and $\sigma_{2}$ are the standard deviations of the two samples.

From our analysis we observe that the pitch patterns are higher in non-empathic segments. The spectral features such as centroid and flux are more stable and smooth when the agent is empathic compared to abrupt changes in non-empatic segments. Spectral patterns captures the perceptual impression of sharpness of sounds. The vocal pattern of loudness is higher in non-empathic situations while it is low when the agent is empathic. We also observed that the auditory-spectrum bands of non-empathic segments were comparatively higher. Our analysis on the relevance of pitch and loudness for empathy signal realization is consistent with the findings of \cite{weiste2014prosody} and \cite{xiao2014modeling}. There are no significant differences in the relevance of features in Table~\ref{table:feat-analysis} for male and female speakers.

The purpose of this analysis was to understand which type of feature could be used for designing classification model. We later observed that our analysis corroborates with the finding of the feature selection procedures in classification experiments. 

\begin{table}[!t]
\caption{Statistical significance (two-tailed two-sample t-test) of acoustic features is reported for each category ( rows ). In the second and third column we report largest \textit{p-value} and the range of effect size (Cohen's \textit{d}) for female and male speakers, respectively. The value of d, $~0.2$, $~0.5$ and $~0.8$ denotes small, medium and large effect sizes, respectively \cite{cohen1988statistical}. The t-value provides the t-statistic of the t-test. The greater, i.e., either positive or negative, the value of the t-statistic, the greater the evidence against the null hypothesis. The closer to zero  the more likely that there is no significant difference.}
\label{table:feat-analysis}
\centering

\begin{tabular}{|l|l|l|}
\hline
\textbf{Feature type} & \multicolumn{1}{c|}{\textbf{Female}} & \multicolumn{1}{c|}{\textbf{Male}} \\ \hline
\multirow{2}{*}{
	\pbox{5cm}{Pitch (F0, voice-probability, voice-quality)}} & p\textless=6.04E-06 & p\textless=6.73E-03 \\ 
& d=[0.3 to 1.1] & d=[0.2 to 1.5] \\
& t=[-3.5 to 14.0] & t=[-2.2 to 15.0]\\[1ex] \hline

\multirow{2}{*}{Loudness} & p=3.08E-25 & p=1.86E-29 \\ 
& d=0.7 & d=1.1 \\
& t=9.1 & t=10.9 \\[1ex] \hline

\multirow{2}{*}{Zcr} & p=4.81E-08 & p=2.01E-05 \\ 
& d=0.3 & d=0.4 \\
& t=-3.6 & t=-3.3 \\[1ex] \hline

\multirow{5}{*}{
	\pbox{5cm}{Spectral (Energy in bands: 0-650 Hz, 250-650Hz, 1-4kHz; Roll-off points 25\%, 50\%, 75\%, 90\%; Position of spectral maximum and minimum, Centroid, Flux)}} & p\textless=5.66E-03 & p\textless=3.64E-04 \\ 
& d=[0.2 to 1.4] & d=[0.2 to 1.8] \\
& t=[-17.6 to 17.6] & t=[-15.0 to 16.8] \\[7ex] \hline

\multirow{3}{*}{\pbox{4cm}{Auditory-spectrum bands 0-25 for male and 0-21 for female}} & p\textless=6.25E-04 & p\textless=1.26E-04 \\  
& d=[0.2 to 0.7] & d=[0.3 to 1.0] \\
& t=[1.1 to 8.9] & d=[2.9 to 9.6] \\ \hline

\end{tabular}
\end{table}

\subsection{Lexical Feature Analysis} %%%%%%%% Morena %%%%%%%%%%
\label{ssec:lexical_feature_analysis}
%\textcolor{red}{TO DO: need clarification for the example of "let us see"}

Several categories of personnel who interact with customers or patients, including call center agents and physicians, are trained to improve their communication skills and develop empathy in their interactions through  careful choice of words \cite{coulehan2001let, Barrett1996}. For example, they are recommended to use phrases such as \textit{``I understand''} when they are listening to the customer who is explaining their problem, or to use \textit{``I would''}. For example, after hearing a customer's story, agent may respond by saying \textit{``I would be upset as well if I were in a similar situation''}, before proceeding to propose possible solutions or provide advice. 

In our call center corpus, we analyzed the lexical realization occuring in empathic and neutral segments by comparing the different word frequencies and POS distributions of unigrams, bigrams and trigrams respectively.  The goal was to extract  the most relevant word $n$-grams from empathic as well as neutral segments. 

We tested the statistical significance over the observed differences with a {\em two-tailed two-sample t-test} and a p-value of $0.01$ with the same number of conversations ($n=302$) we used for the analysis of acoustic features. 

The comparison between neutral and empathy word trigrams showed that agents' phrases such as {\em vediamo un po'} ({\em let's see}), {\em vediamo un attimo} ({\em let's see a bit}), {\em vediamo subito allora} ({\em let's see now, then}) are statistically significant lexical cues for the agents while interacting and manifesting their empathy. It is worth noticing that the Italian verb {\em vedere} ({\em see}) is more frequently used in the first person plural; the same holds true for other frequent verbs such as {\em facciamo} ({\em let's do}) and {\em controlliamo} ({\em let's check}). Those lexical choices are usual when the speaker cares about the problem of the other person. Similarly, significantly different rankings in the empathic distribution affect unigrams, bigrams and trigrams such as {\em non si preoccupi} ({\em do not worry}), {\em allora vediamo} ({\em so, let's see}) that are often used in Italian to grab the floor of the conversation and reassure the other person. For the above lexical features p-value was $<0.01$ and the range of effect sizes, d, was $[0.7~to~3.8]$, and t-statistic was $[-5.5~to~-3.0]$ as shown in Table \ref{table:lex_feature_analysis}.

\begin{table}[h]
\centering
\caption{Statistical significance (two-tailed two-sample t-test) of lexical features. p: p-value of t-test, d: effect size, t: t-stat.}
\label{table:lex_feature_analysis}
\begin{tabular}{|l|l|l|l|}
\hline
\rowcolor{Gray}
\multicolumn{1}{|c|}{\textbf{Lexical Features}} & \multicolumn{1}{c|}{\textbf{p}} & \multicolumn{1}{c|}{\textbf{d}} & \multicolumn{1}{c|}{\textbf{t}} \\ \hline
vediamo un po \textit{(let's see)} & 1.6E-03 & 1.0 & -3.3 \\ \hline
vediamo un attimino (\textit{let's see a bit)} & 2.7E-03 & 1.7 & -3.6 \\ \hline
vediamo subito allora \textit{(let's see now, then)} & 9.7E-03 & 3.8 & -4.6 \\ \hline
vedere \textit{(see)} & 3.3E-05 & 1.1 & -4.4 \\ \hline
facciamo \textit{(let's do)} & 6.1E-07 & 1.3 & -5.5 \\ \hline
controlliamo \textit{(let's check)} & 1.2E-03 & 0.7 & -3.3 \\ \hline
non si preoccupi \textit{(do not worry)} & 5.0E-03 & 1.1 & -3.0 \\ \hline
allora vediamo \textit{(so, let's see)} & 4.2E-04 & 0.9 & -3.7 \\ \hline
assolutamente \textit{(absolutely)} & 4.1E-06 & 1.8 & -5.4 \\ \hline
\end{tabular}

\end{table}

Regarding the POS distributions, the Adverbs that occur frequently in the empathy distribution, such as {\em assolutamente} ({\em absolutely}) and {\em perfettamente} ({\em perfectly}), may have a kind of evocative potential for showing understanding of the other person's point of view, in particular when they are uttered with a tone of voice appropriate to the context.

\section{Automatic Classification of Empathy} %%%%%%%%%%% Firoj %%%%%%%%
\label{sec:classification}
In this Section, we describe the training of an automatic classification system for the recognition of empathy from spoken conversations. We report experimental details of the feature extraction, fusion and classification task and discuss the results. 

\subsection{Classification Task and Data-set} 
\label{ssec:classification_task}
In the automatic classification experiment, our goal was to investigate the empathic conversational segments of the operator.  We have selected a subset of the corpus, that includes a total of 526 conversations annotated with automatic speech transcriptions as well as as {\tt Neutral, Empathy} segment labels. This subset of the corpus allows us to perform a complete computational model training and evaluation in noisy and clean input signal conditions. We partitioned the data-set into train, development and test with 70\%, 15\% and 15\% partitions and no-speaker overlap amongst them. In order to train and evaluate the system we extracted neutral-empathic segment-pair from these conversations, which has been obtained from manual annotation. 
\begin{table}[ht]
\centering
\caption{Segment duration statistics (seconds)  of the {\tt Neutral} and {\tt Empathy}  segment-pairs  and the total amount of speech for each category (hours).}
\label{table:dur-stat}
\scalebox{0.95}{
	\begin{tabular}{|l|r|r|r|r|r|}
	\hline
	\multicolumn{1}{|c|}{\textbf{Class}} & \multicolumn{1}{c|}{\textbf{Avg. (s) }} & \multicolumn{1}{c|}{\textbf{Std. (s)}} & \multicolumn{1}{c|}{\textbf{Total (h)}} & \multicolumn{1}{c|}{\textbf{\# of Seg}} \\ \hline
	\tt{Empathy} & 19 & 13 & 3 & 526  \\ \hline
	\tt{Neutral} & 220 & 148 & 32 & 526 \\ \hline
	\end{tabular}
}
\end{table}

Duration-based descriptive statistics of these segment pairs are provided in Table \ref{table:dur-stat} along with averages and standard deviations on the natural distribution of the data. The segment length of neutral is comparatively longer than the empathic segment as we see in the Table, since it spans from the start of the conversation until the onset of the first empathic event. The total net duration of these segment-pairs is $\sim$35 hours. It is quite usual that in real-world conversations, the distribution of neutral segments is significantly higher than the manifested emotions as can also be seen in \cite{batliner2010segmenting}.

\subsection{Classification System}
\label{ssec:classification_system}
\begin{figure}[!t]
\centering
\includegraphics[width=2.5in]{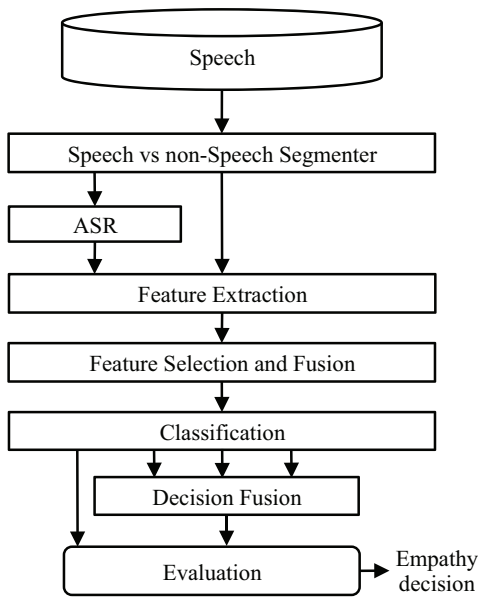}
\caption{The empathy classification system. The pipeline is used both for training and testing the performance of the task.}
\label{fig:classifier}
%\vspace*{-4mm}
\end{figure}

In Figure \ref{fig:classifier}, we present a computational architecture of the automatic classification system, which takes the agent's speech channel as input, then pass it to the automatic speech \textit{vs} non-speech segmenter. After that it generates a binary decision for each speech segment of the agent's behavior in terms of neutral \textit{vs} empathy. 
%The recognition system evaluates the cues present in each speech segment then commits to the binary decision. 
In order to evaluate the relative impact of lexical features we considered the case of noisy transcriptions (left branch in Figure \ref{fig:classifier}) provided by an automatic speech recognizer (ASR). We extracted, combined, and selected acoustic features directly from the speech signal and generated the classifier's training set. We implemented both feature fusion and decision fusion algorithms (bottom part of~Figure \ref{fig:classifier}) to investigate the performance of different classifier configurations. This architectural design can be used in real time application, which combines all automatic processes. 
%In the next sections we describe the building blocks of the classification system in details.

\subsection{Speech \textit{vs} Non-Speech  Segmenter}
\label{ssec:automatic-segmenter}

An HMM-based speech \textit{vs} non-speech segmenter has been trained using a set of $150$ conversations, containing approximately $100$ hours of spoken content and used Kaldi \cite{povey2011kaldi} for the training and decoding processes. Training data has been prepared using force-aligned transcriptions. Mel Frequency Cepstral Coefficient (MFCC) and their derivatives has been used as features. Number of gaussian and beam width have been optimized using a development set of $50$ conversations. Final model has been tested using a test of $50$ conversations. The F-measure of the system was $66.0\%$ on the test set.

\subsection{Undersampling and Oversampling}
\label{ssec:under-oversampling}

The statistics in Table \ref{table:dur-stat} manifest the data imbalance problem for the two classes, {\tt Empathy} and {\tt Neutral}. Once the manual segments are processed through the automatic segmenter, the ratio of {\tt Empathy}/{\tt Neutral} labeled segments  is $6\%$ {\em vs} $94\%$. A common approach to cope with that is via oversampling or undersampling in the data or feature space. We have under-sampled the instance of the majority class (i.e., neutral) at the data preparation phase and oversampled the minority class (i.e., empathy) after the feature extraction phase. In the literature it is reported that the combination of oversampling and undersampling often leads to better performance \cite{chawla2002smote}.  

The idea of our undersampling is a novel approach, in which we randomly pick samples from the majority class. At the same time, we maintain a variation of the random samples. The intuition behind is that classifier does not learn much if it basically sees many similar samples. Our way of picking random samples maintains a variation and upon feeding such samples for the same class classifier learn better predictors. We maintain this variation by defining a set of bins with different segment lengths, and then randomly select $N$ segments from each bin.  
We used $N=1$ for this study. The number of bin and size of $N$ has been optimized empirically on the development set and by investigating the descriptive statistics such as percentiles, mean and standard deviation. The pseudocode of the algorithm is presented in Alam et el. work \cite{alam2017affectivescene}. The implementation of this algorithm will be made publicly available. The undersampling stage generated a $18\%$ {\em vs } $82\%$ ratio of 
{\tt Empathy} {\em vs } {\tt Neutral} segments up from the initial $6\%$ {\em vs} $94\%$.

For oversampling we used Synthetic Minority Oversampling Technique (SMOTE) \cite{chawla2002smote} and its open-source implementation in Weka \cite{witten2005data}. In SMOTE, the upsampled examples (i.e., empathy) are generated based on the $K$ nearest neighbors of the minority class. Nearest neighbors have been chosen randomly based on the percentage of target oversampling. It computes the difference between the feature vector and its nearest neighbor. Then, multiply this difference by a random number between 0 and 1 and add it to the feature vector. 
%More details of this approach can be found in \cite{chawla2002smote}. 
The oversampling was tuned on the development set and we achieved a further improvement on the imbalance problem. Before oversampling the class distribution  was $18\%$ {\em vs} $82\%$, and following oversampling that it became $30\%$ {\em vs} $70\%$. Using this approach classifier learns the variations of segments in different lengths and also reducing the effect of imbalance class distribution. 

\subsection{Feature Extraction}
\label{ssec:feature_extraction}
In this section we report the algorithms used in the extraction of the acoustic, linguistic and psycholinguistic features in the empathy recognition task. We discuss differences and similarities of the features used in our experiments to those used in other emotion and personality recognition tasks \cite{schuller2013paralinguistics, alam2014icassp}. 

\subsubsection{Acoustic Features}
\label{sssec:acoustic_features}
Recent studies in emotion and personality recognition showed that the paralinguistic properties of speech are well represented by low-level features~\cite{schuller2009acoustic, schuller2013paralinguistics, alam2014icassp}. We followed a similar approach and we extracted a very large number of low-level features and their statistical functionals using the openSMILE tool~\cite{eyben2013recent}. 
The low-level acoustic features include the feature set of the computational paralinguistic challenge's (COMPARE-2013) feature set \cite{schuller2013interspeech}, Geneva minimalistic acoustic feature set \cite{eyben2015geneva} and few more formant features. For the sake of the replicability we made the configuration files of the feature set publicly available\footnote{https://github.com/firojalam/openSMILE-configuration}. 
We extracted low-level acoustic features at approximately $100$ frames per second. Regarding voice-quality features the frame size was $60$ milliseconds with a Gaussian window function and $\sigma=0.4$. Regarding other low-level features the frame size was $25$ milliseconds with a hamming window function. The details of the low-level features and their statistical functional are provided in Table \ref{table:acoustic_features}. After feature extraction, the size of the resulted feature set is $6861$.

\begin{table}[h]
\caption{Extracted low-level acoustic features}
\label{table:acoustic_features}
\centering	
\scalebox{0.7}{
\begin{tabular}{l}
\hline
\multicolumn{1}{c}{\textbf{Low-level acoustic features}}   \\ \hline
\rowcolor{DarkGray}\textbf{Voice Quality}\\ \hline
Probability of voicing, jitter-local, jitter-DDP, shimmer-local,\\
log harmonics-to-noise ratio (HNR)\\ \hline
\rowcolor{DarkGray}\textbf{Cepstral}\\ \hline
MFCC 1-14\\ \hline
\rowcolor{DarkGray}\textbf{Spectral}\\ \hline
Auditory spectrum (RASTA-style) bands 0-25 (0-8kHz),\\
Spectral energy 250-650Hz, 1-4kHz,\\
Spectral roll-off points (0.25, 0.50, 0.75, 0.90),\\
Spectral slope 0-500Hz and 500-1500Hz,\\
Spectral flux, centroid, entropy, variance, skewness, kurtosis,\\ Spectral slope,\\
Difference of spectral flux between two consecutive frames\\
Psychoacoustic spectral sharpness, harmonicity\\
Alpha Ratio - ratio of the total energy from 50-1000Hz, and 1-5 Hz\\		
Relative energy of formant 1, 2 and 3 \\
First and second harmonic difference\\
First and third harmonic difference\\
Hammarberg index\\
\hline
\rowcolor{DarkGray}\textbf{Prosody}\\ \hline
F0 final, F0 envelope, F0final with non-zero frames, \\
Root-mean-square signal frame energy, \\
Sum of RASTA-style auditory spectra,\\
Loudness, Zero crossing rate,\\
Formant frequencies [1-4], bandwidths [1-4] \\\hline
\hline
\end{tabular}
}
\end{table}

\begin{table}[h]	
\caption{Statistical functionals}
\label{table:acoustic_features2}
\centering	
\scalebox{0.7}{
	\begin{tabular}{l}
	\hline
	\multicolumn{1}{c}{\textbf{Low-level acoustic features}}   \\ \hline
	\rowcolor{DarkGray}\textbf{Voice Quality}\\ \hline
\multicolumn{1}{c}{\textbf{Statistical functionals}}  \\\hline
Percentile 1\%, 99\% and percentile range 1\%-99\%   \\
Quartile (1-3) and inter-quartile (1-2, 2-3, 3-1) ranges   \\
Relative position of max, min, mean and range  \\
Arithmatic mean, root quadratic mean   \\
Mean of non-zero values (nnz)  \\
Contour centroid, flatness
Std. deviation, skewness, kurtosis   \\
Uplevel time 25, 50, 75, 90, \\
Rise time, fall time, left curvature time, duration   \\
Mean, max, min and Std. deviation of segment length   \\
Linear prediction coefficients (lpc0-5), lpc-gain   \\
Linear regression coefficients (1-2) and error   \\
Quadratic regression coefficients (1-3) and error  \\\hline

\end{tabular}
}
\end{table}

\subsubsection{Lexical Features}
\label{sssec:lexical_features}
We extracted lexical features from automatic transcriptions. The automatic transcriptions were generated using a large vocabulary ASR system \cite{chowdhuryunsupervised2014}. We designed HMM-based ASR system using a subset of $1894$ conversations containing approximately $100$ hours of spoken content and a lexicon of size $\sim$15K words. From the conversations we extracted Mel Frequency Cepstral Coefficient (MFCC) features and then spliced by taking 3 frames from each side of the current frame. It was followed by Linear Discriminant Analysis (LDA) and Maximum Likelihood Linear Transform (MLLT) feature-space transformations to reduce the feature space. Then, we trained acoustic model using speaker adaptive training (SAT). In order to achieve the best accuracy we also used Maximum Mutual Information (MMI). The Word Error Rate (WER) of the ASR system was $31.78\%$ on the test set and $20.87\%$ on the training set, using a trigram language model of perplexity $87.69$. For the training and decoding process, we used Kaldi \cite{povey2011kaldi}.

The training set of the ASR system overlapped with the task classification test corpus that was used in the classification experiments (Section~\ref{ssec:classification_evaluation}). However, the training error rate of $20.87\%$ was still realistic and useful for the classification task.

We mapped the speech transcriptions into lexical features in the step preceding the training of the classifier. The transcriptions of each segment were converted to bag-of-words vectors weighted with logarithmic term frequencies (tf) multiplied with inverse document frequencies (idf), presented in equation \ref{eq:tf-idf}. 
\begin{equation}
\label{eq:tf-idf}
\scalebox{.95}{
%\resizebox{\linewidth}{!}{
	$tf\times idf=\\log(1+f_{ij}) \times log\left ( \frac{\text {number of segments}}{\text {number of segments that include word i}} \right ) $
}
\end{equation} 
where $f_{ij}$ is the frequency for word $i$ in conversation $j$.
In order to take advantage of the contextual benefits of $n$-grams, we extracted trigram features. Because this resulted in an unreasonably large dictionary and we filtered out lower frequency features by preserving 10K most frequent $n$-grams.

\subsubsection{Psycholinguistic Features}
\label{sssec:psycholinguistic_features}
Similar to the lexical features we extracted the so-called psycholinguistic features from automatic transcriptions. Over the past few decades, Pennebaker et al. have designed psycholinguistic word categories using high frequency words and developed the Linguistic Inquiry Word Count (LIWC) \cite{pennebaker2001linguistic}. These word categories are mostly used to study gender, age, personality, and health to estimate the correlation between these attributes and word usage (see \cite{alam2014icassp,morena2015} and the references therein). The types of LIWC features include the following: 

\begin{enumerate}
\item General features such as word count, average number of words per sentence, percentage of words found in the dictionary and percentage of words longer than six letters and numerals. 
\item Linguistic features include pronouns and articles.
\item Psychological features include affect, cognition and biological phenomena. 
\item Features about personal concern include work and home.
\item Paralinguistic features include accents, fillers and disfluencies.
\item Punctuation categories include periods and commas. 
\end{enumerate}

We used the dictionary that is available within LIWC for Italian \cite{Alparone2004}. The Italian dictionary contains $85$ word categories. LIWC extracted $5$ general descriptors and 12 punctuation categories constituting a total of $102$ features.  The LIWC feature processing differs according to types of features. Some features are counts and others are relative frequencies (see \cite{tausczik2010psychological}).   

\subsection{Feature Fusion and Selection}
\label{ssec:feature_combination_selection}
%Following the feature extraction step in Figure~\ref{fig:classifier},
We applied feature selection to each individual feature set (Figure~\ref{fig:classifier}). We also merged acoustic and lexical features into a single vector to represent each instance in a high-dimensional feature space. Let $A=\left \{ a_{1},a_{2},...,a_{m} \right \}$ and $L=\left \{ l_{1},l_{2},...,l_{n} \right \}$ denote the acoustic and lexical feature vectors respectively. The feature-combined vector was $Z=\left \{ a_{1},a_{2},...,a_{m}, l_{1},l_{2},...,l_{n} \right \}$ with  $Z\in R^{m+n}$. Given the high-dimension of the feature vector and the {\em curse of dimensionality}, we applied feature selection on the $Z$ space to achieve an optimal feature dimension of size $k$, $k< \left (m+n\right)$. Another objective of the feature selection process was to find the best compromise between the dimension of the input and the performance of  a target classifier.

For feature selection we used the {\em Relief} \cite{kononenko1994estimating} feature selection technique. In  \cite{alam2013comparative}, we comparatively evaluated the Relief method against other algorithms and it outperformed them in classification performance and computational cost. We ranked the feature set according to Relief scores and generated learning curves by incrementally adding batches of ranked features. We then selected the optimal set of features by stopping when performance saturated or started decreasing~\cite{alam2013comparative}. 

The goal of Relief is to estimate weights to find relevant attributes with the ability to differentiate between instances of different classes under the assumption that nearest instances of the same class have the same feature values and different class have different values. Relief estimates weight of a feature, $F$, using Equation \ref{eq:relief}. \\
\begin{equation}
\label{eq:relief}
W[F]=P(x|nearest\;miss) - P(x|nearest\;hit)
\end{equation}

where $x$ is a value of $F$, \textit{nearest miss} and \textit{nearest hit} are the nearest instances of the same and a different class, respectively.

Since some feature selection algorithms do not support numeric feature values such as information gain and suffer from data sparseness such as Relief, we discretized feature values into $10$ equal-frequency bins \cite{witten2005data} as a pre-processing step of feature selection. The equal-frequency binning approach divides data into $k=10$ groups, where each group contains approximately equal number of values. Moreover, the feature selection and discretization approach were performed on development set (see Section \ref{ssec:classification_evaluation}) in order to avoid biases in classification experiments. The equal-frequency binning approach and the size of the bin, $k=10$, were empirically found optimal in other paralinguistic classification task \cite{alam2013comparative}. Furthermore, we did not apply the selection algorithm to the psycholinguistic features due to the limited size of the feature set.

After selecting the features, we analyzed the top-ranked features per category. The top ranked acoustic features included spectral, mfcc, probability of voicing, and pitch. The spectral feature type included spectral variance, flux, auditory-spectrum with bands $1K-4K$ Hz, roll-off points.
The highly relevant LIWC feature category included verb, article, negations, and, social and cognitive processes. The lexical features included \textit{allora vediamo se} (so let's see if), \textit{assolutamente} (absolutely), \textit{sicuramente} (certainly) \textit{tranquillamente} (nicely) and \textit{non si preoccupi} (do not worry). The statistical analysis of acoustic and lexical features for the neutral-empathy segment annotation (see Section \ref{ssec:acoustic_feature_analysis} and \ref{ssec:lexical_feature_analysis}) corroborate these latter findings of the top-ranked features designed for the classification step.

\subsection{Classification and Evaluation}
\label{ssec:classification_evaluation}

\subsubsection{Baseline}
\label{sssec:baseline}
To compare the results of our proposed approach we have conducted an experiment to have baseline results. The typical approach of computing baseline to choose majority class \cite{witten2005data} or random selection of class labels. For our study, we have chosen the latter approach, to compute the baseline, we have randomly selected the class labels based on the prior class distribution. 

\subsubsection{Proposed Classification Models}
\label{sssec:proposed_models}
%\textcolor{red}{TO DO: Understanding the challenges...}

We designed our classification models using Support Vector Machines (SVMs). One important challenge in designing classification models using acoustic, lexical and combination of $acoustic+lexical$ features was their higher dimensional feature space compared to the size of the dataset. In order to deal with such a problem, we have chosen to use the linear kernel of the SVMs, which we found useful in other paralinguistic task \cite{alam2013comparative}. For designing the classification model using psycholinguistic features we used the Gaussian kernel of the SVMs as the dimensionality of this feature set is small compared to other sets. For the experiment, we used Sequential Minimal Optimization (SMO) \cite{Platt1998}, which is designed to solve quadratic optimization problems of SVMs. This is available with an open-source implementation by Weka machine learning toolkit \cite{witten2005data}.  
For the training, we optimized the penalty parameter $C$ of the error term by tuning it in the range $C\in [10^{-5},...,10]$ and the Gaussian kernel parameter $G$ in the same range as well, using the development set. To obtain the results on the test set we combined the training and development set and trained the models using the optimized parameters. We have not attempted to use a deep neural network, due to the smaller sized data-set.

Regarding the classifiers trained on different feature sets, we combined decisions from each classifier by applying {\em majority voting}, as in Equation \ref{eq:majority_voting}.  
\begin{equation}
\label{eq:majority_voting}
H(s)=c_{\hat{j}};\;\; 
where\;\hat{j} =argmax_{j}\sum_{i=1}^{T} h_{i}^{j}(s)
\end{equation}

where $H(s)$ is the combined classification decision for an input instance $s$; $h_{i}^{j}(s)$ is the output of the classifier $h_{i}$ for the class label $c_{j}$; $i=1...T$ is the number of classifiers; $j=1...C$ is the number of classes. 

In the experiments we used SMO across different classification systems and evaluated different sets of features such as lexical {\em vs} acoustic. 
%After applying the speech non-speech segmenter we removed the non-speech segments, which resulted the total spoken content to $\sim$23 hours from $\sim$35 hours. 

\subsubsection{Evaluation}
\label{sssec:evaluation}
\begin{figure}[ht]
\centering
\includegraphics[width=4in]{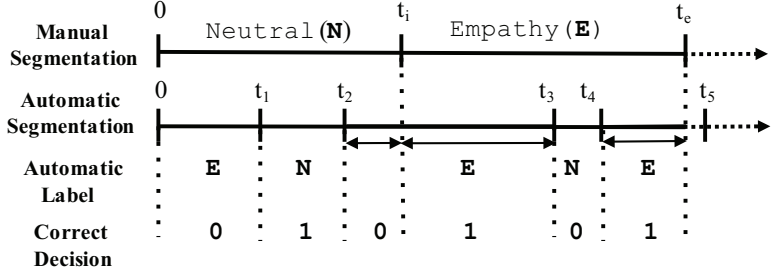}
\caption{Sample alignment  of the manual and automatic segments and their labels. The evaluation spans from $t=0$ to $t=t_e$, the end of the {\tt Empathy} segment. Automatic labels are evaluated against the reference labels and error statistic are computed (last row).}
\label{fig:alignment}
\end{figure}

Evaluation at the segment level poses a great challenge. The evaluation procedure requires an alignment of the manual segment and labels with the output of the automatic segmentation and classification system described in the previous section. In Figure~\ref{fig:alignment}, we present an alignment example of the reference (manual) and automatically generated segments and their labels. The reference segmentation spans from $t=0$ to $t=t_e$ and labels the {\tt Neutral} ({\textbf {\tt N}}) segment spanning from $t=0$ to $t=t_i$ and the {\tt Empathy} ({\textbf {\tt E}}) segment from $t=t_i$ to $t=t_e$. 
Automatic segments inherit the reference label that falls inside its boundaries (e.g., the segment spanning the interval [$0,t_1$] or [$t_3,t_4$]).
For the evaluation purpose, automatic segments that span across the onset, $t=t_i$, and end, $t=t_e$,  (e.g., the segment spanning the interval [$t_2,t_3$]) are split in two segments with two distinct reference labels. For instance the segment spanning [$t_2,t_3$] will be evaluated with the segment [$t_2,t_i$] (reference label ({\textbf {\tt N}})) and the segment [$t_i,t_3$] (reference label ({\textbf {\tt E}})). The alignment process will generate the correct label statistics for all segments as shown on the last row of Figure~\ref{fig:alignment}. 

We have measured the performance of the system using the Un-weighted Average (UA), which has been widely used in the evaluation of paralinguistic tasks \cite{schuller2009interspeech}. We have extended such measure to account for the segmentation errors as evaluated in similar cases by NIST in diarization tasks \cite{nist-rt2009,castan2014audio}. UA is the average recall of positive and negative classes and is computed as $UA=\frac{1}{2}\left (\frac{tp}{tp+fn} +\frac{tn}{tn+fp} \right)$, where $tp$, $tn$, $fp$, $fn$ are the number of true positives, true negatives, false positives and false negatives, respectively. 
We computed $tp$, $tn$, $fp$ and $fn$ from a weighted confusion matrix, as shown in Equation \ref{eq:confusion_matrix}, where the weight for each instance is the corresponding segment length: 

\begin{equation}
\label{eq:confusion_matrix}
C(f)=\left \{ c_{i,j}(f)=\sum_{s \in S_T}\left [ ((y=i)\wedge (f(s)=j))\times length(s) \right ] \right \}
\end{equation}

In Equation \ref{eq:confusion_matrix}, $C(f)$ is the $2 \times 2$ confusion matrix of the classifier $f$,  $s$ is the automatic segment in the test set $S_T$, including the segment with boundary $t_i$ and $t_e$ (see Figure~\ref{fig:alignment}), $length(s)$ is the duration of $s$, $y$ is the reference label of $s$, $f(s)$ is the automatic label for $s$. The indices $i$ and $j$ represent the reference and automatic class label of the confusion matrix.

\subsection{Classification Results}
\label{ssec:results_discussion}

\begin{table}[!t]
\renewcommand{\arraystretch}{1.3}
\caption{Empathy classification results at  the segment level using acoustic, lexical, and psycholinguistic (LIWC) features together with feature and decision level fusion.  Maj - Majority voting. Dim. - Feature dimension.}

\label{table:results_classifier}
\centering
\begin{tabular}{|l|r|r|}
\hline
\rowcolor{Gray}
\textbf{Experiments} & \textbf{Dim.} & \textbf{Test-Set}\\ \hline
\textbf{Random baseline} & & 49.3 \\ \hline
\textbf{Acoustic} & 2400 & 68.1 \\ \hline
\textbf{Lexical} & 8000 & 65.6 \\ \hline
\textbf{LIWC} & 89 & 67.3 \\ \hline
\textbf{Acoustic+Lexical} & 2600 & 68.3 \\ \hline
\textbf{Maj(Acoustic+Lexical+LIWC)} &  & 70.1 \\ \hline
\end{tabular}
\end{table}

In Table \ref{table:results_classifier}, we report the performances of the automatic classification system trained on different feature types: lexical (automatic transcriptions), acoustic and psycholinguistic. We report test set results for feature combination-based system as well as and classifier combination. In the latter system we applied majority voting.  
For the statistical significance, we have computed McNemar's significant test over the test set \cite{mcnemar1947note}.
 
For single feature-type systems, acoustic-based models provided the best performance compared to lexical and psycholinguistic alone. The results of acoustic-based system are significantly better than random baseline with $p<2.2E-16$. The acoustic-based system provides a useful and low-computation  classification model, when no automatic transcriptions are available. LIWC's system performance improve over the lexical-only system with very few lexical features ($89$). In addition, all system's UAs are higher and statistically significant compared to the baseline results with $p<2.2E-16$, chi-square value $228.6~to~1866.9$ and d $0.4~to~1.1$. For McNemar test, the test statistic is usually computed by a paired version of Chi-square and effect size, d, is computed using Cramer's $\phi$. The value of $\phi$, $0.1$, $0.3$ and $0.5$ represents small, medium and large effect respectively.   

In terms of feature and system combination, we obtained the best results with \textit{majority voting}. The statistical significance test showed that the results of the majority voting are statistically significant compared to any other system's results with $p<=0.0004$, chi-square value $12.2~to~1446.5$ and d $0.1~to~1.0$. Compared to the baseline, the best model for automatic classification provides a relative improvement over the baseline of $35.7\%$. 
Linear combination of lexical with acoustic features has not improved performance, despite its success in other paralinguistic tasks \cite{alam2014icassp}. It has not even improved performance when combined with feature selection.

In order to assess the impact of the automated transcriptions, in a different study, we compared the performance between automatic and manual transcriptions for a automatic classification of emotional state. The results show that performance differences are very low, only $1.2\%$ drop with automated transcriptions \cite{alam2016empathy}. Therefore, we found that the use of automatic transcriptions are reasonable for the experiment given that manual transcriptions are not available in all cases.

\subsection{Discussion}
\label{ssec:limitations_future_work}

The ability to detect empathy is very useful in all those human tasks that are geared towards the assessment of human interactions (e.g., teacher-student, doctor-patient, customer-operator etc.). 

This paper presents the results of a quantitative study of empathy in spoken conversations. In order to obtain such results, we annotated a corpus of real human-human spoken conversations by using an annotation framework that was defined for this work and based on the psychological `modal model of emotions' by Gross. The goal of the human annotators was to label the speech segments where they could perceive a possible expression of empathy. Their perception task was guided by the operational definition of empathy that was based on the  `modal model' theory. The novelty of our emotion annotation framework is related to the fact that it provides the annotators with operational definitions of the target emotions (in this case empathy) that takes into consideration the situational context where the emotions unfold. 

This annotation task was done on a continuous time scale: this is a difficult task due to the variability of emotional expressions and to the individual nature of the perceptual process of human beings. Also, the identification of the emotional segment boundary is a difficult task, and reaching an agreement on the same boundary position in continuous time scale is a further challenge. Providing the annotators with operational definitions of emotions, such the one we provided for empathy in the present work, was useful for constraining those individual variability factors, and the results obtained by evaluating the inter-annotator agreement may support this view.

We have extensively analyzed acoustic and lexical features to understand their correlations with empathy. We observed that certain features are highly correlated with empathy. For designing the automatic classification system, we used acoustic, lexical and psycholinguistic features.  
From the classification results in Section \ref{ssec:results_discussion}, we observe that the results of our proposed classification model perform better than random baseline, which indicates that this results could be used as a starting point. In the comparison of different feature sets, we obtained higher results using acoustic features, and it would be useful when no transcription available. Lexical features rely on ASR system and improvement could be possible with a higher accuracy of the ASR system. LIWC features extracted using transcriptions are based on the lexical mapping, which can also be improved by enriching the lexical knowledge. One possible approach of enriching lexical knowledge is by using Wordnet or finding similar words from word-embedding vector \cite{mikolov2013efficient}. 

The motivation of the study of feature level combination was that emotional aspects are represented in both verbal and vocal-nonverbal acoustic space and upon combining them in linear space can provide better results. However, the challenge here is that the vector space representations are different such as dense representation in acoustic space and sparse representation of lexical space. Future studies might need to focus finding a better combination strategy. One possible approach would be transforming the sparse lexical representation into dense representation either using word-embedding vector or other feature transformation approaches such as principal component analysis.

The present work suggests that it is possible to detect empathy. From our analysis, we observed that there are spoken language features that correlate well with empathy. The pitch and loudness are amongst the most correlated acoustic features ( see table~\ref{table:feat-analysis} ) with empathy. The lexical features also play an important role in signaling empathy to the conversation participant and we have shown statistical significant patterns from the Italian corpus (see table~\ref{table:lex_feature_analysis}. 

There are aspects of the empathy that are very relevant and could benefit from the annotation framework and models presented in this paper. In particular the aspect of the coordination of emotions (e.g., anger or frustration) on one side (e.g., customer) and empathy on the other side (e.g., agent side). We have focused on analyzing behavioral cues both in terms of paralinguistic and linguistic cues, which would be useful in designing empathic virtual agent. For example, lexical content ``let us see, perfect'' with medium pitch voice could be useful cues to represent empathic response.These aspects are very important if we need to design predictive models for human-machine conversations or automatically evaluate the interactions of individuals. 

One of the important research challenges is the variability of empathy in diverse affective scenes, domains, cultures, and languages. We believe the annotation model and classification system are robust enough to be extended or adapted to such different contexts to create new baselines and work on adaptation models such as~\cite{daume2009frustratingly,blitzer2007biographies,ganin2016domain} to further enhance the model. 

Another important contribution of this paper is the automatic classification system of empathy in spoken conversations. Its design and classification results indicate that it is possible to support in the future a controlled-model in a {\em empathic} human-computer dialogue system. For human-machine scenario, further study is needed to understand that the coordination between a virtual empathic agent and a human can achieve a stable equilibrium. The analysis we have carried out maybe a starting point to train this coordination model, however further human-machine experimental studies will be needed.

\section{Conclusion}
\label{sec:conclusion}
Empathy refers to an emotional state triggered by a shared emotional experience. In this paper, we have addressed the problem of observing and annotating instances of empathy in real-life spoken conversations. The annotation process describes the scene through the Situation-Attention-Appraisal-Response model. We have operationalized the definition of empathy and designed an annotation process of the human-human dialogues in call centers. We have designed and evaluated a system that automatically segments and classifies empathic events from spoken conversations. We have investigated the effectiveness of acoustic, lexical, and psycholinguistic features and their combinations. 
The performance of our best classification system is very promising and significantly above the random baseline. The annotation model and classifier performances may lead to extensions in other situations and applications in human-machine interaction and automatic behavior analysis. In future work, we foresee to extend this work with a comparative study of the concurrent signals from agent and operator channels and understand the effect of interlocutor's behaviors on each other. The empathic behavioral patterns might be different for human-machine interaction scenarios, however, we believe our findings would be helpful to study such interactions in future.

\section*{Acknowledgment}
The research leading to these results has received funding from the European Union - Seventh Framework Programme (FP7/2007-2013) under grant agreement n\degree{} 610916- SENSEI.

\section*{References}

\bibliography{./bibliography_merge_v3,acl2016}

\end{document}